# XDANNG: XML based Distributed Artificial Neural Network with Globus Toolkit

Hamidreza Mahini
ELearning Department
Iran University of Science and Technology
Tehran, Iran
h_mahini@vu.iust.ac.ir

Alireza Mahini
Academic Staff of Computer Engineering Department
Islamic Azad University-Gorgan branch
Gorgan, Iran
mahini@comp.iust.ac.ir

Javad Ghofrani
Computer Department
Iran University of Science and Technology-Behshahr Branch
Behshahr, Iran
Javad.Ghofrani@yahoo.com

*Abstract*— **Artificial Neural Network is one of the most common AI application fields. This field has direct and indirect usages most sciences. The main goal of ANN is to imitate biological neural networks for solving scientific problems. But the level of parallelism is the main problem of ANN systems in comparison with biological systems. To solve this problem, we have offered a XML-based framework for implementing ANN on the Globus Toolkit Platform. Globus Toolkit is well known management software for multipurpose Grids. Using the Grid for simulating the neuron network will lead to a high degree of parallelism in the implementation of ANN. We have used the XML for improving flexibility and scalability in our framework.**

## I. INTRODUCTION

With the emergence of AI and utilization of ANN, with the imitation from nature a new era in the creation of a new artificial intelligence was introduced. In 1957, by creating the first neural model, Perceptron, some scientists believed that they have found the key to make an intelligent being with the power of learning equal to human beings.

ANN had to face massive amounts of processing and data, while there was no advanced instrument to simulate the complicated behavior of a natural neural network like the human being's brain or even a simpler intelligent mechanism, the legend of creating a competitive intelligent system with the human's brain fade out. During the last two decades of the 20th century, some scientists started to create test beds to obtain some of the advantages of the ANN, such as the high degree of learning and recognition. These researches have been continued up to now.

With the daily improvements in calculation and communication technologies, this belief that mankind could have a wide network for implementing an artificial brain improves more and more. In this regard, some efforts have occurred in the various fields such as hardware, software, ANN patterns and also ANN learning methods. Nowadays we use the ability of ANN for different kinds of calculation and pattern recognition fields but the implementations of ANN are in a low level of throughput and have some restrictions in scalability.

After 1995 some efforts for using the computational power of supercomputers in some US universities, named I-WAY project, led to Grid born. Using the computation and storage of computers in all over the world for massive processes was the purpose of this project.

Computers are designed in the way that they can perform any action right after another in high speeds such as million operations just in a second. Although the human brain contains more active details, it works with lower speed. This issue is the main difference between computers and the human brain. Computers can perform sequential operations fast and precise, but in the parallel arena human brain works better. We can clarify this issue with an example; if a human wants to calculate the summation of two hundred digit numbers, the suggested result will take a considerable time and at the end the answer might not be true. Hence, computers can perform this kind of operations fast and exact. Because of the sequential origin of summation, electrical mechanism of CPU performs faster than chemical mechanism of human brain. However the human brain has a wonderful speed in recognizing an image.

As long as the human brain was presented as a dynamic system with parallel structure and different processing ways in comparison with other usual methods, researches and interests about ANN commenced. At the 1911, efforts for perception of brain mechanism strengthened particularly when Segal announced that brain is made up of main parts, named neurons. Human brain as a data processing system is a parallel system formed by more than 100 trillion neurons which are connected together with $10^{16}$ connections.

Perhaps the first efforts in this regard go back to the late 19<sup>th</sup> century and the beginning of the 20<sup>th</sup> century, when basic works on the physics, psychology and neurophysiology were carried out which are the prerequisites for the artificial neural network. These studies where done by scholars such as "Hermann von Helmholtz", "Ernst mach" and "Ivan Pavlov" but never led to such a mathematical model from "Neuron". In the 4<sup>th</sup> decade of the 20<sup>th</sup> century, when "Warren McCulloch" and "Walter Pits" found that neural networks can evaluate every logical and calculus functions, a new horizon of ANN was commenced [1]. However the first practical usage of ANN was emerged by introducing the Perceptron by "Frank Rosenblatt" in the 5<sup>th</sup> decade of 20<sup>th</sup> century (1958). After a while, in 1960 "Bernard Widrow" introduced the comparative linear Adaline neural network, which was similar to Perceptron network in the aspect of its structure, in addition with a new learning principle. But the restrictions of both Adaline and Perceptron networks were that they both could only categorize patterns which had linear differences. ANN used to promote until the 7<sup>th</sup> decade of the 20<sup>th</sup> when in the 1972 "teuvo kohonen" and "James Anderson" introduced a new ANN which were able to act as storage elements while in the same decade "Stephen Grosssberg" was working on the self organizer networks. In the 6<sup>th</sup> decade of the 20<sup>th</sup> century because of absence of new ideas and the lack of sophisticated computers for implementation of Neural Networks, activities on Neural Networks where slowed down. But during the 8<sup>th</sup> decade with emerging new ideas and ascending growth in the technology of microprocessors, studies about ANN where also increased significantly.

The probable reason for popularity of neural networks in future could be the high speed of computers and algorithms which will ease the usage of neural networks in industrial aspects.

In this study we try to suggest a basis for implantation of ANN which could be compared with biological neural.

## II. INTRODUCTION OF GRID COMPUTING

Gird Computing in simple word is a new revolution and transmutation for distributed computing. The goal of Grid Computing is to represent a massive virtual parallel computer with the power of collection of sharing resources of heterogeneous systems.

Defining new standards of resource sharing, concurrent with availability of higher bandwidths were two major causes for big steps toward Grid technology [3].

### A. Globus Project

The Globus project is a multi-institutional research effort that seeks to enable the construction of computational grids providing pervasive, dependable, and consistent access to high-performance computational resources, despite geographical distribution of both resources and users.

The organization of these activities consists of four main parts:

- Research
- Software tools
- Testbed
- Software

In this study we used Globus Toolkit as a simple software tool for installation of computational Girds and Grid-based software.

A prominent element of the Globus system is the Globus Toolkit, which defines basic services and capabilities required to construct a computational grid.

Globus toolkit is an open source and open architecture project. Some other important features of Globus Toolkit are as below:

- Security: authentication, authorization, safe data transition
- Resource management: management and submission of remote jobs
- Information services: query about resources and status of them

### III. IMPLEMENTED WORKS FOR ANN SIMULATION

Despite of huge growth in performance of processors and processing networks, still a lot of calculation problems such as performance resistance, floating point, and memory bandwidth have remained unsolved because of the lack of main resources. For example the problems of aerology, biology, and physic of big energies are included in this area.

Using supercomputers to solve such problems just supplies one or two levels of parallelism order. Nowadays by emerging fast processors and calculation systems, the lack of computational resources (even supercomputers) and hard management of HPC systems are big barriers in the way of solving such parallel problems.

Under this circumstance and current problems the idea of grid computing to strengthen and integrate the computational power for solving such problems with wide and massive calculations was emerged.

In other way, attempts to create facsimiles of these biological systems electronically have resulted in the creation of artificial neural networks. Similar to their biological counterparts, artificial neural networks are massively parallel systems capable of learning and making generalizations. The inherent parallelism in the network allows for a distributed software implementation of the artificial neural network, causing the network to learn and operate in parallel, theoretically resulting in a performance improvement.

In this part we will take a look at some efforts that tried to link parallel computational systems to ANN. These attempts can be good patterns for further projects. Considered experiences in these researches are so valuable in science and applications.

### A. Implementation of ANN on the Cactus parallel platform

This project was done by Center for Computation and Technology Louisiana State University. In this study, implementation of a Neural Network on the Cactus framework was considered and at the end the advantages and disadvantages of this system and expected further improvements were discussed.

The idea of this paper is that 'the main area where performance improvements are expected are within the neuron calculations'. As far as the calculation of Neural Network originally has a high level of parallelism, each neuron performs these calculations and the calculations are independent of the calculations performed by the other neurons on its layer. Additionally, the simplicity of these calculations allow an entire layer to be generalized in a single operation: taking the dot product of the weights column vector with the input vector and passing the result to the output function. These results in simplification of neuron calculations move from a nested loop to a single loop.

As such, layers can be split into discrete 'clusters' of neurons residing on the same physical processor. After each layer completes its calculations, all the clusters need to pass the updated weights, bias and output back to the simulation. The computations could then continue for the next layer (or take whatever action would be most appropriate)

### Initial Serial Pseudo Code for Calculations:

```
for(x=0;x<OUTPUTS;x++)
    for(y=0;y<INPUTS;y++)
        value+=weights[x][y]*inputs[y];
    output[x]=outfunc(value);
```

### Parallel Pseudo Code for Calculations:

```
for(x=0;x<OUTPUTS;x++)
    value[x]=outfunc(weights[x] <DOT> input);
```

(DOT represents the parallel multiply)

The learning functions of an ANN are somewhat harder to implement in parallel. While the calculations in a neuron involve simple operations, learning requires adjusting the weight matrix.

The parallel ANN performed somewhat worse than the single processor network in all the tests, although the actual performance difference did decrease with larger inputs and entries (as was expected). There are several possible explanations for the performance degradation. One possible explanation is that the MPI overhead was substantially greater than expected. It should also be noted that PETSc is aimed to solve partial differential equations in parallel. It is possible that coding the MPI by hand instead of using PETSc for MPI would improve performance.

This idea could not meet the expected results and scientists behind this idea mentioned some reasons such as correlation between performance and the amount of input values and also the overhead of intercommunication between processors. We have suggested some solution for these two problems in the final part of our study.

### B. N2Grid Research Project

N2Grid Research Project was done by Erich Schikuta and his team members at Vienna University in Austria. The suggested approach of their research was to emphasis the usage of new emerging Gird technology as a transparent environment that allows users to exchange information (neural network objects, neural network paradigms) and exploit the available computing resources for neural network specific tasks leading to a Grid based, world-wide distributed neural network simulation system.

It implements a highly sophisticated connectionist problem solution environment within a Knowledge Grid. Their system uses only standard protocols and services in service oriented architecture, aiming for a wide dissemination of this Grid application. In this project the Grid was used as a middleware of resource sharing that helps to intercommunicate among different virtual organizations. The grid services are used for accessing remote resources of grid for simulations and sharing of models and patterns of ANN among users [6]. The main point of this research was to use the Gird services, while the potential capability of Grid as a huge parallel computing system was ignored.

N2Grid is an evolution of the existing NeuroWeb system that was born by the N2grid system creators. The idea of this system was to see all components of an artificial neural network as data objects in a database.

It aims for using the Internet as a transparent environment to allow users the exchange information (neural network objects, neural network paradigms) and to exploit the available computing resources for specific tasks related to Neural Network (specifically training of neural networks). The principles of NeuroWeb design are: acceptance, homogeneity and efficiency [7].

*C. An implementation of neural networks framework for parallel training using PVM*

This research was an attempt to implement the executive parallel platform for implementing the ANN by a cluster of IBM-PC machines. An optimized object-oriented framework to train neural networks, developed in C++, is part of the system presented. A shared memory framework was implemented to improve the training phase. One of the advantages of this system is its low cost, considering that its performance can be compared to similar powerful parallel machines.

The PVM message passing interface in the Linux operating system environment is used for this project.

The restrictions of Cluster systems such as maintenance costs, lack of scalability and restriction of resources affects the performance of this idea. But with the parallel feature of ANN, this system is a good solution for simulating tiny and simple processing Neural Networks.

*D. NeuroGrid*

The idea behind NeuroGrid is to provide a framework for finding information within a distributed environment. This project is used the distributed platform of web for acting like a network of neurons. This idea is based on the power of biological neuron networks in searching. by this approach we can implement a search and data access system with distributed nodes via the Internet. When a user sends a request to find a subject, this request will be sent to the other subnets and the search will continue along the parallel subnets. With this method we can have a distributed search steam that imitates from neural networks and decreases the time and complexity of infrastructure of massive search engines [9].

*E. Represented hardware models*

Development of digital neural hardware is extracted as a response to need for speedup in ANN simulations, and the purpose of these hardware is to obtain a better cost to performance ratio in despite of general purposed systems.

The main approach in speedup of ANN algorithms is the parallelization of processes. With this method, some strategies were developed for efficient mapping of neural algorithms to parallel hardware. Due to diversity of neurohardware, choosing the optimal hardware platform for implementing an ANN algorithm seems hard, therefore some kind of hardware design classification would allow a better evaluation of existing designs.

These issues are used to classify Digital neurohardware:

- System architecture
- Degree of parallelism
- Typical neural network partition per processor
- Inter-processor communication network
- Numerical representation

According to the system type and the architecture of neural network that we want to use, we must choose one of the digital hardware. However this may help to speedup the neural network simulation, but decreases the level of flexibility and scalability.

Among the efforts which have done in this field we can consider CNAPS (product of adaptive solutions) and SYNAPSE from Siemens, these are samples of neural computers and using the special purposed neural chips.

Among the challenges that digital neurohardware faces today, the competition with general-purpose hardware is probably the toughest one: computer architecture is a highly competitive domain which advances at an incredible pace. On the other hand, the area of ANN hardware is not yet as commercialized as general-purpose hardware. Also digital neurohardware tends to be more algorithms specific. This requires a good knowledge about algorithms as well as system design and leads to a high time-to-market. Therefore, the general-purpose computers can profit more often from advances in technology and architectural revisions. In addition, general-purpose hardware seem to be more user-friendly and therefore offer higher flexibility than single-purposed hardware (like neurohardware).

## IV. IMPLEMENTATION OF ANN ON THE GRID INFRASTRUCTURE

As far as ANN systems are complicated and parallel massive systems, they need a wide and parallel infrastructure for efficient performance. Supplying such an infrastructure is beyond the power of PCs and supercomputers and in most cases of the implementation of ANN it was via colonization or neurons hardware.

On the other hand, with the emergence of new grid technology and parallel usage of individual systems which are widespread around the world we can propose this hypothesis: if we map all the nodes of ANN system to available computational resources around the world, it would be more feasible to achieve the goal of real ANN power.

Because of the lack of power in implementation of the real ANN and observation of enormous processing power of this artificial intelligence, this technology has not attracted enough attentions. But recently with the increasing power of computer systems and at the same time, developments in their speed, reliability of computer networks it would be more probable that in the coming years, we can achieve a fast and powerful network infrastructure and a worldwide Grid. With the existence of such a system we can observe the real power of ANN.

## V. ARCHITECTURE OF XDANNG FRAMEWORK

Our XDANNG framework is made up of four main parts which are its building block units, that let it perform in mentioned way. These four parts are as below:

## A. XML Parser:

This unit is responsible for reading and parsing the XML formatted input files of framework. It reads the XML nodes of inputted files and compares it with declared patterns in section 8 (of this paper) to find the mismatches occurred in the structure of input files.

## B. Interpreter:

The interpreter unit is responsible for reading the structure of network from the input files and creating the related nodes of ANN .The input of this unit is prepared from XML Parser Unit. In other words, this unit makes out the requested structure of ANN that will be simulated by XDANNG.

## C. MU (Mapping Unit):

This unit is responsible for distributing and mapping the generated nodes to the Grid nodes. For improving the runtime performance, this unit can decide which node of ANN is suitable for running on which node of grid. The Assigner unit uses the provided Globus Toolkit standard services to do its duty.

## D. Execution Manager:

the execution manager is in charge of coordination of other three units and also it is the manager of simulation execution of the XDANNG in bootstrapping, execution and output phases. This duty will occur with feeding input layer, and summation of output results from output layer of simulated ANN.

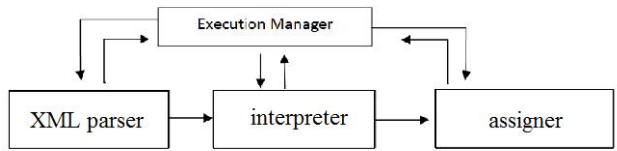

Figure 1: XDANNG Framework building blocks

## VII. INTEGRATION WITH GLOBUS TOOLKIT

Our XDANNG framework is introduced as an upper layer for Globus Toolkit middleware. With this integration we can use the distributed power of underlying fabric.

We used layered model for integration of XDANNG with Globus Toolkit. This method causes the higher level of abstraction and a better flexibility. The mentioned model is illustrated below:

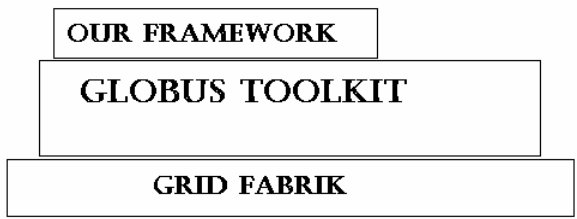

Figure 2: Integration with Globus Toolkit

In Figure 2 there are 3 layers. The first and lower layer is the grid infrastructure. It consists of many heterogeneous systems and networks. As described before the Globus Toolkit is a middleware to use computational girds and also here is used to manage this great base system. At the top of these 2 layers stays our XDANNG system that uses the Globus Toolkit services for implementing neural networks simulation.

## VIII. USAGE OF XML FOR DECLARING THE STRUCTURE AND OPERATION OF NEURAL NETWORK

In this study we analyze a simple ANN system which is made up of N nodes and l layers. To strengthen the flexibility, we used XML which gives us a unique format to define architecture of neural network and current connections between nodes.

According to this format, inputs, network architecture and weight values matrix, and inputs of each node would be available in the XML formatted files. Finally we can define the output nodes in the XML format too.

For instance, we use the below structure for representing network architecture:

```xml
<architecture>
  <node>
    <nodeIndex>...</nodeIndex>
    <preNodes>
      <element>
        <nodeIndex>...</nodeIndex>
        <inputIndex>...</inputIndex>
      </element>
      ...
      <element>
        <nodeIndex>...</nodeIndex>
        <inputIndex>...</inputIndex>
      </element>
    </preNodes>
    <b>...</b>
    <function>...</function>
  </node>
  ...
  <node>
    ...
  </node>
</architecture>
```

Figure 3: Representing network architecture

As it has been mentioned before, ANN is an originally parallel system, therefore requires a parallel programming for its implementation in our research project. In this regard we shall use the MPI standard library which has implemented in C and Fortran Languages. Although there are other tools available, but because of the high speed of written programs in the C language we will use this language for implementing our framework.

SMPD is an appropriate model which is supported by MPI standard library for Grid programming. In this model a program would be run on all nodes but we can make each node to run only a part of the program and process related data to that node. To transfer a program to its destination nodes we can use the file transfer services of Globus Toolkit.

Used SPMD Programming template for our project is demonstrated as below:

```
Main ()

{

        Initialize

        Common part of code

        If (current process==i)

        {        Do specified job for node i

        }

        Else if (current process== i+j)

        {

                Do specified job for node i+j

        }

        ...

        ...

}//end of main
```

For instance, imagine the below neural network:

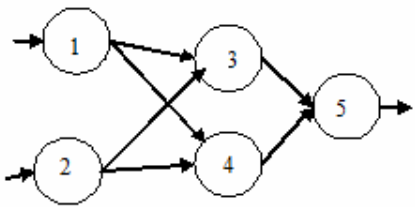

Figure 4: Sample Neural Network

To define the architecture of this model network in our platform we use such a below file:

```xml
<architecture>
    <node>
        <nodeIndex>1</nodeIndex>
        <b>e</b>
        <function>f(x)=x</function>
    </node>
```

```xml
<node>
    <nodeIndex>2</nodeIndex>
    <preNodes>
        <element>
            <nodeIndex>1</nodeIndex>
            <inputIndex>0</inputIndex>
        </element>
        <element>
            <nodeIndex>3</nodeIndex>
            <inputIndex>1</inputIndex>
        </element>
    </preNodes>
    <b>e</b>
    <function>f(x)=x</function>
</node>
<node>
    <nodeIndex>3</nodeIndex>
    <b>e</b>
    <function>f(x)=x</function>
</node>
<node>
    <nodeIndex>4</nodeIndex>
    <preNodes>
        <element>
            <nodeIndex>1</nodeIndex>
            <inputIndex>0</inputIndex>
        </element>
        <element>
            <nodeIndex>3</nodeIndex>
            <inputIndex>1</inputIndex>
        </element>
    </preNodes>
    <b>e</b>
    <function>f(x)=x</function>
</node>
<node>
    <nodeIndex>5</nodeIndex>
    <preNodes>
        <element>
            <nodeIndex>2</nodeIndex>
            <inputIndex>0</inputIndex>
        </element>
        <element>
            <nodeIndex>4</nodeIndex>
            <inputIndex>1</inputIndex>
        </element>
    </preNodes>
    <b>e</b>
    <function>f(x)=x</function>
</node>
</architecture>
```

To clarify the above text we should take into consideration that:

The tag <node> is used to illustrate nodes of neural network. To define the connection between nodes we should note that: (1) the outputs of which node would be used as inputs of the node i. (2) the outputs of node j would be which input in node i.

To make this capacity in neural network system architecture we have used the tag <preNode>. This tag contains some <element> tags. Every signal of these <element> tags illustrates one of the inputs of our slightly nodes. For example the node with index 2 in the above architecture takes its inputs from the node1 and 3. With this method we can increase the flexibility of our proposed structure. The tag <b> defines the value of bias. Finally <function> tag will define the transition function.

To define the input values of neural network the below schema is used:

```
<node>
    <nodeIndex>i</nodeIndex>
    <items>
        <item>
            <inputIndex>j</inputIndex>
            <value>k</value>
        </item>
        ...
    </items>
</node>
```

Figure 5: Neural Network input nodes

To illustrate the output nodes of whole neural network we use the same method, i.e. similar structures to above structure are used to define the final nodes which are the results neural network process will be exported.

For the above example we have an input structure like:

```
<inputValues>
    <node>
        <nodeIndex>1</nodeIndex>
        <items>
            <item>
                <inputIndex>0</inputIndex>
                <value>i</value>
            </item>
            <nodeIndex>3</nodeIndex>
            <items>
                <item>
                    <inputIndex>0</inputIndex>
                    <value>k</value>
                </item>
            </items>
        </items>
    </node>
</inputValues>
```

Figure 6: Input Structure sample

Other values of neural network such as weight values of each node are also definable for our proposed platform.

## IX. Conclusion

The first part of this paper gave an overview of artificial neural networks. A classification of existing works would help to identify different approaches in implementations of neural networks. To make efficient use of the artificial neural networks, we suggested the new emerging Grid infrastructure and also we mentioned main issues to design a framework to implement artificial neural networks on the Gird test bed using MPI standard library and XML.

With the suggested framework, if we would have a Grid of computers of a country or computers in all over the world, we can obtain a power beyond the power of human biological neural networks. In our model if the number of nodes approaches to the infinite, we can ignore the communications delay because of the independent calculation of each layer from other layers during the process time of nodes of that layer.

In other hand ,with the daily growth of computer networks in speed and ma of data processed and transmitted via computer networks, achieving the real power of artificial neural networks will not be an out of reach goal.

Advantage of our model are its independency to the model and pattern of implemented Neural network, flexibility, scalability and also using of well-known and famous Globus Toolkit as its infrastructure to help us to hide the details of distribution and use of its services.

Using functions of MPI standard library will take to account as one of delays in speeding up of this model because of their layer model of implementation that makes big decrease in runtime.

## X. Future Work

We suggest solving some NP-hard and NP-complete problems with XDANNG and this method can be compared with some other general methods. But implementing this suggestion needs using million nodes to create a powerful XDANNG.

At least, implementing such a system for simulating ANN with dynamic structure on the Globus Toolkit platform can be suggested.

REFERENCES

[1] Ahmar Abbas. *Grid computing: A Practical Guide to Technology and Applications*, Charles River Media Inc, 2004

[2] R. Beal & T. Jackson, *Neural Computing: An Introduction*, Institute of Physics Publishing, 1998**.**

[3] IBM Redbooks, *Introduction to Grid Computing with Globus*, second Edition, September 2003.

[4] IBM Redbooks, *Enabling Applications for Grid Computing with Globus*, 2003.

[5] Ian Wesley-Smith, "A Parallel Artificial Neural Network Implementation", Center for Computation and Technology, Louisiana State University Baton Rouge, Louisiana, 70803 Faculty Adviser: Dr. Gabrielle Allen ,Charles River Media © 2004

[6] Schikuta Erich, Weishäupl Thomas, "N2Grid: Neural Networks in the Grid", International Joint Conference on Neural Networks (IJCNN), Budapest, Hungary, July 2004.

[7] Schikuta,E, "NeuroWeb: an Internet-based neural network simulator", Inst. fur Informatik und Wirtschaftsinformatik, Univ. of Vienna, Austria, 2004.

[8] Araújo M.A.A., Teixeira E.P., Camargo F.R., Almeida J.P.V., Pinheiro H.F., "An Implementation on Neural Networks Framework for Parallel Training using PVM and Shared Memory", IEEE Congress , Vol.2, No.1315 – 1322, 2003

[9] Stanford K.Boahen "Neurogrid: Emulating a Million Neurons in the Cortex", 28th Annual International Conference of the IEEE, 2006.

[10] Schoenauer T., Jahnke A.,Roth U., Klar H., "Digital Neurohardware; Principles and Perspectives", Institute of Microelectronics, Technical University of Berlin,1998

AUTHORS PROFILE

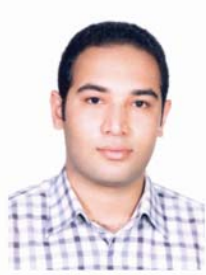

**Hamidreza Mahini,** received the B.S. degree in software engineering in 2006 from Iran University of Science and Technology (IUST), Iran. Now he is a M.S. degree candidate in IT engineering in the same university. His current researches focus on grid computing, semistructured data models such as XML, Artificial Neural Network, and high performance IP route lookup.

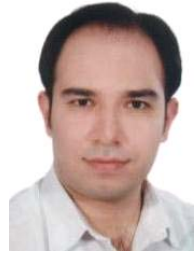

**Alireza Mahini,** received the Master degree in computer systems architecture in 2006 from Iran University of Science and Technology (IUST), Iran. Now he is a Ph.D candidate in the Islamic Azad University, sciences and researches branch. He is academic staff of Islamic Azad University, Gorgan branch and president of Gorgan SAMA college. His current researches focus on distributed systems, high performance IP route lookup, network processor architecture, and NOC.

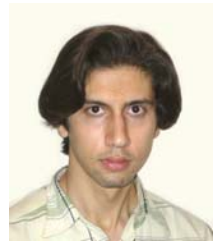

**Javad Ghofrani,** is a B.S. student in software engineering in Iran University of Science and Technology (IUST), Behshahr branch. His current researches focus on grid computing, MPI, and semistructured data models such as XML.